\documentclass{article}

\usepackage[preprint,nonatbib]{neurips_2026}

\usepackage{graphicx}
\usepackage{amsmath}
\usepackage{amssymb}
\usepackage{amsfonts}
\usepackage{booktabs}
\usepackage{multirow}
\usepackage{colortbl}
\usepackage{xcolor}
\usepackage{float}
\usepackage{microtype}
\usepackage{xspace}
\usepackage{wrapfig}
\usepackage[normalem]{ulem}

\usepackage{caption}
\usepackage[accsupp]{axessibility}

\usepackage{hyperref}

\usepackage{algorithm}
\usepackage{algorithmic}

\usepackage[capitalize]{cleveref}
\crefname{section}{Sec.}{Secs.}
\Crefname{section}{Section}{Sections}
\Crefname{table}{Table}{Tables}
\crefname{table}{Tab.}{Tabs.}
\crefname{figure}{Fig.}{Figs.}
\Crefname{figure}{Figure}{Figures}

\usepackage{orcidlink}

\newcommand{\eg}{\emph{e.g.}\xspace}
\newcommand{\ie}{\emph{i.e.}\xspace}

\renewcommand{\acksection}{\section*{Acknowledgment}}

\definecolor{mylightgray}{RGB}{255,255,255}
\definecolor{mygreen}{RGB}{224,233,236}
\definecolor{myblue}{RGB}{236,243,221}
\definecolor{best}{RGB}{212, 237, 218}
\definecolor{best2}{RGB}{214, 228, 244}
\definecolor{mygreen}{RGB}{34,139,34}
\definecolor{mylightblue}{RGB}{0,162,230}
\definecolor{myyellow}{RGB}{255, 215, 0}
\definecolor{customPurple}{RGB}{104,52,154}

\newcommand{\model}{\mathcal{F}}
\newcommand{\bc}{\mathbf{c}}

\newcommand{\modelname}{\textsc{UniGP}\xspace}
\newcommand{\methodname}{\textsc{DUGP}\xspace}

\begin{document}

\title{\textsc{UniGP}: Taming Diffusion Transformer for Prior-Preserved Unified Generation and Perception}

\author{
\textbf{Qin Guo}$^{1}$ \quad
\textbf{Hao Luo}$^{2,3,4}$ \quad
\textbf{Dongxu Yue}$^{5}$ \quad
\textbf{Weixuan Jin}$^{5}$\\
\textbf{Xiao Fu}$^{6}$ \quad
\textbf{Fan Wang}$^{2}$ \quad
\textbf{Dan Xu}$^{1,7,\dagger}$\\[4pt]
$^{1}$The Hong Kong University of Science and Technology\\
$^{2}$DAMO Academy, Alibaba Group, Zhejiang, China \quad
$^{3}$Hupan Lab, Zhejiang Province\\
$^{4}$Zhejiang University, Zhejiang, China \quad
$^{5}$Tsinghua University\\
$^{6}$The Chinese University of Hong Kong \quad
$^{7}$Zeekr Automobile R\&D Co., Ltd.\\
$^{\dagger}$Corresponding author
}

\maketitle

\begin{center}
    \centering
    \includegraphics[width=0.99\linewidth]{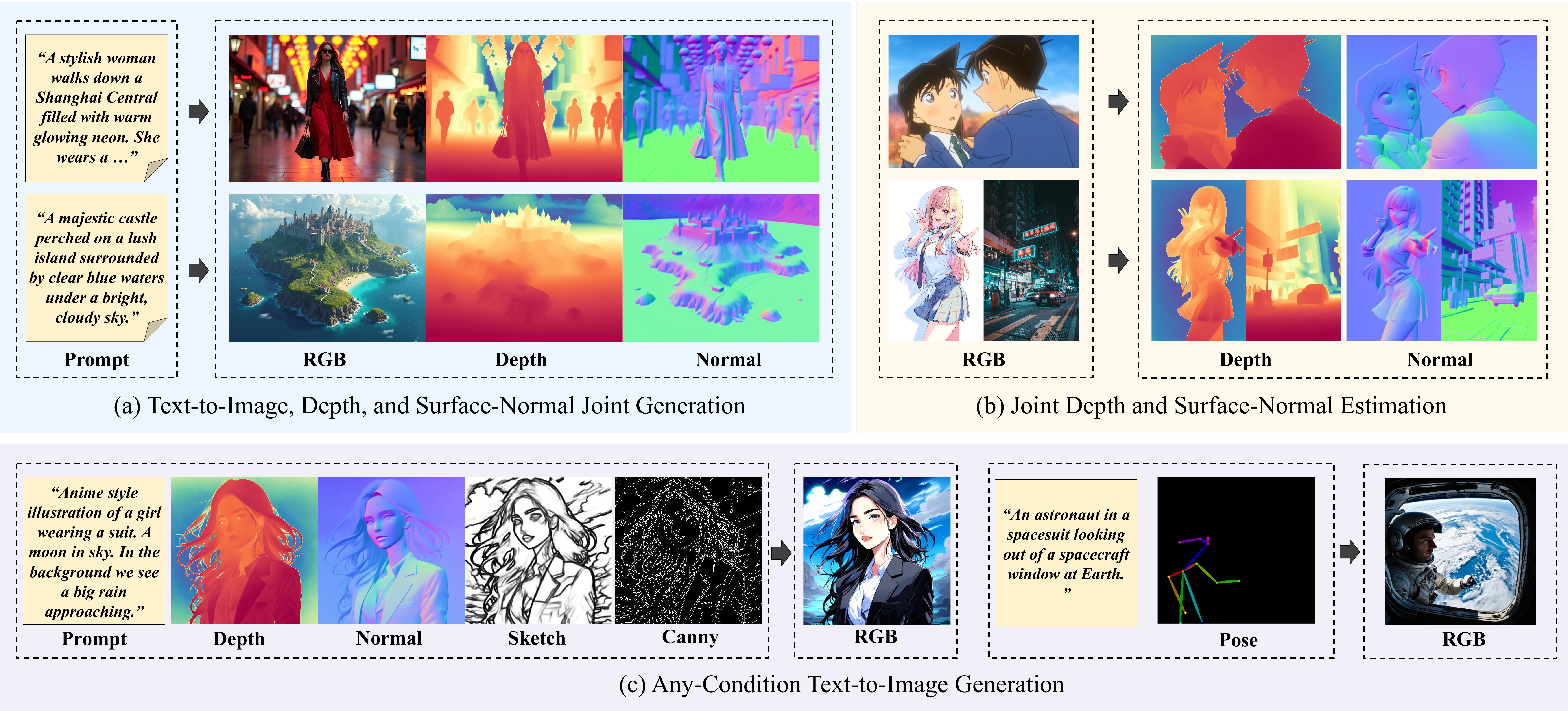}
    \captionof{figure}{We present \modelname, a Diffusion Transformer-based framework that simultaneously models RGB and dense distributions within a single framework, supporting: \textbf{(a)} Text to image, depth, and surface-normal joint generation; \textbf{(b)} Joint depth and surface-normal estimation; and \textbf{(c)} Any-condition text-to-image generation.}
    \label{fig:teaser}
\end{center}

\begin{abstract}
Recent advances in diffusion models have shown impressive performance in controllable image generation and dense prediction tasks. However, existing approaches typically treat diffusion-based controllable generation and dense prediction as separate tasks, overlooking the potential benefits of jointly modeling the heterogeneous distributions.
In this work, we introduce \modelname, a framework built upon MMDiT, which unifies controllable generation and dense prediction through simple joint training, without the need for complex task-specific designs or losses, while preserving the backbone's versatile priors.
By learning controllable generation and prediction under different conditions, our model effectively captures the joint distribution of image-geometry pairs. \modelname~is capable of versatile controllable generation, dense prediction, and joint generation.
Specifically, the proposed \modelname~consists of \methodname~and a unified dataset training strategy. The former, following the principle of Occam's razor, uses only a copied image branch of MMDiT to model dense distributions beyond RGB, while the latter integrates heterogeneous datasets into a unified training framework to jointly model generation and perception tasks.
Extensive experiments demonstrate that our unified model surpasses prior unified approaches and performs on par with specialized methods. Furthermore, we demonstrate that multi-task joint training provides complementary benefits: generative priors enrich perceptual details, while perceptual learning improves structural alignment in generation.
Project page: \href{https://guoqincode.github.io/UniGP/}{guoqincode.github.io/UniGP}.

\end{abstract}

\section{Introduction}
\label{sec:intro}

Large-scale Text-to-Image (T2I) Diffusion models~\cite{rombach2022high,sd3,Midjourney} have recently demonstrated exceptional high-quality image generation. This success has spurred exploration into downstream tasks based on pretrained T2I models, falling into two major categories: 1) Controllable generation, where models like ControlNet~\cite{zhang2023adding} use external conditions (\eg depth, normal) to guide the T2I process. 2) Dense prediction, where models are repurposed for tasks like monocular depth estimation~\cite{ke2023repurposing}, later extended to normal estimation~\cite{ye2024stablenormal} and joint depth-normal estimation~\cite{garcia2024fine,fu2025geowizard}.

Although these methods have achieved remarkable results independently, they only model the transition within a single distribution, limiting them to simple, single-task scenarios.
One might ask: since regression-based feed-forward models already achieve highly accurate dense predictions, why should we pursue perception within a generative framework? The answer lies in the fundamental limitation of regression models: they act as passive, unidirectional observers. While they predict geometry accurately, they cannot internalize this geometric consciousness to actively guide complex creation processes. A truly unified model should embed perception into the generative latent space, allowing geometric understanding to act as a core generative prior.

Previous work, JointNet~\cite{zhang2024jointnet}, aimed to model multiple distributions simultaneously through intricate architectural design. While this approach resulted in a significant increase in parameter count, it achieved suboptimal results. Recent works including UniCon~\cite{li2024unicon}, OneDiffusion~\cite{le2025one}, and JoDi~\cite{xu2025jodi} explored how to model the joint distribution but largely ignored the complementary mechanism between perception and generation tasks, resulting in performance gaps compared to diffusion-based expert models.

In this paper, we propose \modelname, a unified diffusion transformer (DiT) model that learns the joint distribution of different modalities of an image with a flexible architecture.
\textbf{First,} we adopt the Multi-Modal DiT (MMDiT) framework~\cite{sd3}. Following the principle of Occam's razor, we introduce the \textbf{d}isentangled \textbf{u}nified \textbf{g}eneration and \textbf{p}erception branch (\methodname), which only copies the image branch from MMDiT to model distributions beyond RGB, gracefully avoiding the massive computational overhead of replicating the entire backbone.
\textbf{Second,} we propose a unified dataset and training strategy to bridge these heterogeneous tasks. By employing a simple binary loss weighting, our model jointly learns generation and perception, obviating the need for task-specific designs.
\textbf{Third,} we demonstrate the complementary benefits between tasks. Our ablation studies show that jointly training perception and generation within a single framework allows perceptual learning to enforce strict spatial alignment for generation, while generative priors imbue perception with finer details.

Compared to representative controllable generation methods, \modelname~achieves superior controllability with less data. Unlike mainstream dense prediction methods that quickly suffer from catastrophic forgetting, our approach preserves generative capacity, which enhances perceptual generalization.
As shown in~\cref{fig:teaser}, \modelname~supports various tasks within a single model: \textbf{1)} text-to-image, depth and surface-normal joint generation; \textbf{2)} joint depth and surface-normal estimation; and \textbf{3)} any-condition text-to-image generation.

To summarize, our main contributions are three-fold:
\begin{itemize}
    \item We propose \modelname, an MMDiT framework that unifies generation and perception via a \methodname~branch. Following Occam's razor, this branch models non-RGB data by reusing only the backbone's image branch parameters.
    \item We propose a unified dataset and a simple training strategy with a binary loss, enabling joint learning of heterogeneous tasks without task-specific architectures.
    \item Extensive experiments show \modelname~significantly surpasses prior joint models and rivals task-specific experts, demonstrating clear complementary benefits between jointly trained perception and generation tasks.
\end{itemize}

\section{Related Work}
\label{sec:relate}

\noindent\textbf{Controllable Diffusion Models.}
Controllable diffusion models aim to use external conditions to guide generation.
ControlNet~\cite{zhang2023adding} and its subsequent models~\cite{qin2023unicontrol,zhao2024uni,sun2024anycontrol} extend T2I generation by encoding condition signals into latent representations using a trainable UNet encoder, injected into the backbone via zero convolution.
However, these methods are limited strictly to generation tasks and require massive datasets for precise control. In contrast, our method jointly trains controllable generation and dense prediction tasks, enabling faster convergence and superior geometric consistency with significantly less data. We propose a novel controlling module design and training strategy for MMDiT, enhancing capabilities within the current SOTA diffusion framework.

\begin{figure*}[!t]
  \centering
  \includegraphics[width=1.0\linewidth]{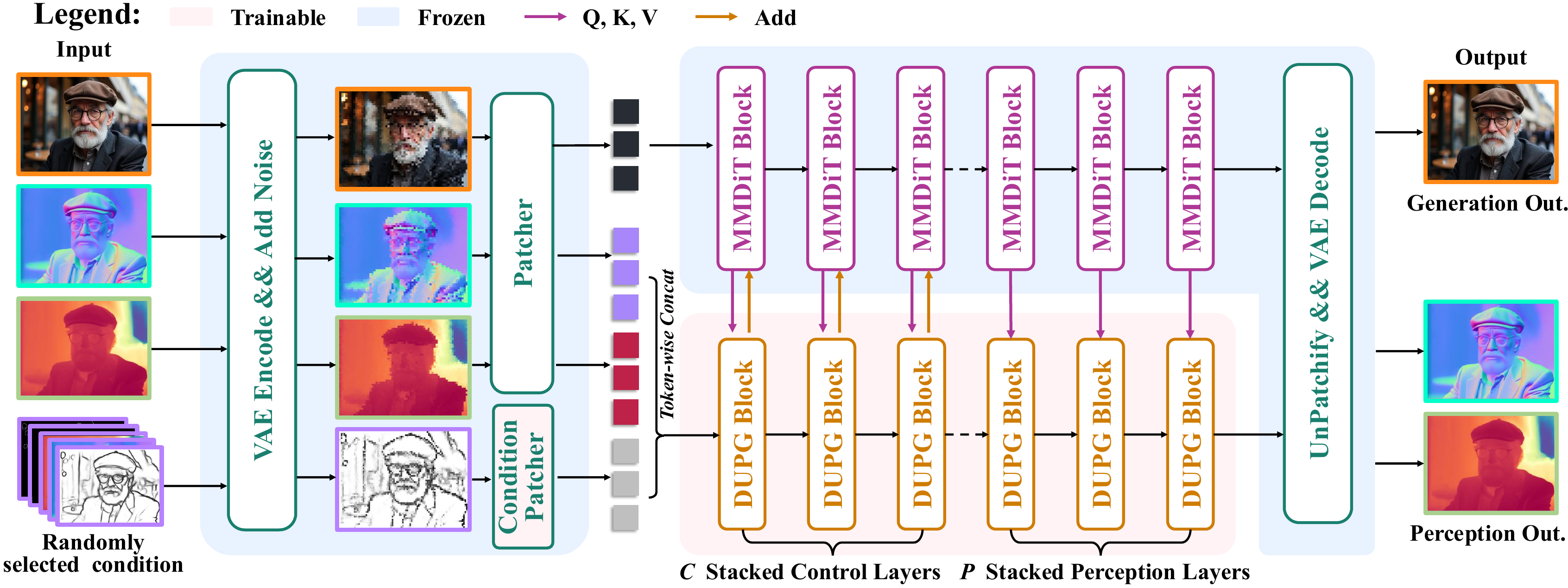}
  \caption{\textbf{Framework of \modelname.}
  \textbf{1)} Our inputs include: \textbf{a)} RGB images; \textbf{b)} Depth and Normal images; and \textbf{c)} Randomly selected condition as described in~\cref{Preliminary}. \textbf{d)} Prompts (omitted for brevity).
  \textbf{2)} After VAE encoding and adding noise, the noisy RGB, depth and normal latents are fed into the backbone's patcher, while the clean condition latents are passed to the \textbf{Condition Patcher} of \methodname. Then, the tokens of noisy depth/normal and condition are concatenated token-wise and pass through the \textbf{Stacked Control Layers} and \textbf{Stacked Perception Layers}.
  \textbf{3)} Finally, the backbone generates RGB images, while \methodname~generates depth and normal maps.
  }
  \label{fig:method:framework}
\end{figure*}

\noindent\textbf{Diffusion Models in Perception Tasks.}
Currently, a notable trend involves adapting pretrained T2I diffusion models for dense prediction tasks~\cite{vandenhende2021multi}, such as depth and surface normal estimation.
Marigold~\cite{ke2023repurposing} fine-tuned Stable Diffusion (SD)~\cite{rombach2022high} to generate precise depth maps, leveraging SD's strong geometric priors.
Follow-up works~\cite{gui2024depthfm,garcia2024fine,he2024lotus,ye2024diffusionmtl,fu2025geowizard,xu2024diffusion} improve its performance and efficiency. StableNormal~\cite{ye2024stablenormal} extends this paradigm to surface normal estimation. GeoWizard~\cite{fu2025geowizard} jointly predicts depth and normal using parallel branches.
The above models entirely fine-tune diffusion for perception tasks, quickly losing versatile generative priors and narrowing their applicability.
Unlike prior works that compromise generation when adapted to perception, our method explicitly bridges generative and perception distributions, quarantining generative priors while learning highly accurate perception.

\noindent\textbf{Unified Diffusion Models.}
While less common, some works have pursued unified diffusion to enable a single model to handle multiple modalities.
LDM3D~\cite{stan2023ldm3d} jointly generates images and depth within an RGBD space.
JointNet~\cite{zhang2024jointnet} adopts a symmetric UNet structure using an inpainting approach to support both generation and perception.
UniCon~\cite{li2024unicon} improves upon JointNet with fewer parameters.
Recent works like OneDiffusion~\cite{le2025one} and JoDi~\cite{xu2025jodi} coarsely model different distributions using Transformer attention before fine-tuning the entire model.
However, previous methods treat generation and perception merely as conditional image generation, largely ignoring the underlying complementary mechanisms between the two tasks, resulting in suboptimal results.
We demonstrate that under a tailored model design and dataset strategy, jointly learning generation and perception yields mutual benefits, achieving state-of-the-art results among unified frameworks.

\section{Method}
\label{sec:method}

To achieve unified generation and perception, we present \modelname. We outline the preliminaries and problem setting in~\cref{Preliminary}, and then introduce \modelname~in~\cref{our_model}.

\begin{figure*}[http]
  \centering
  \includegraphics[width=0.98\linewidth]{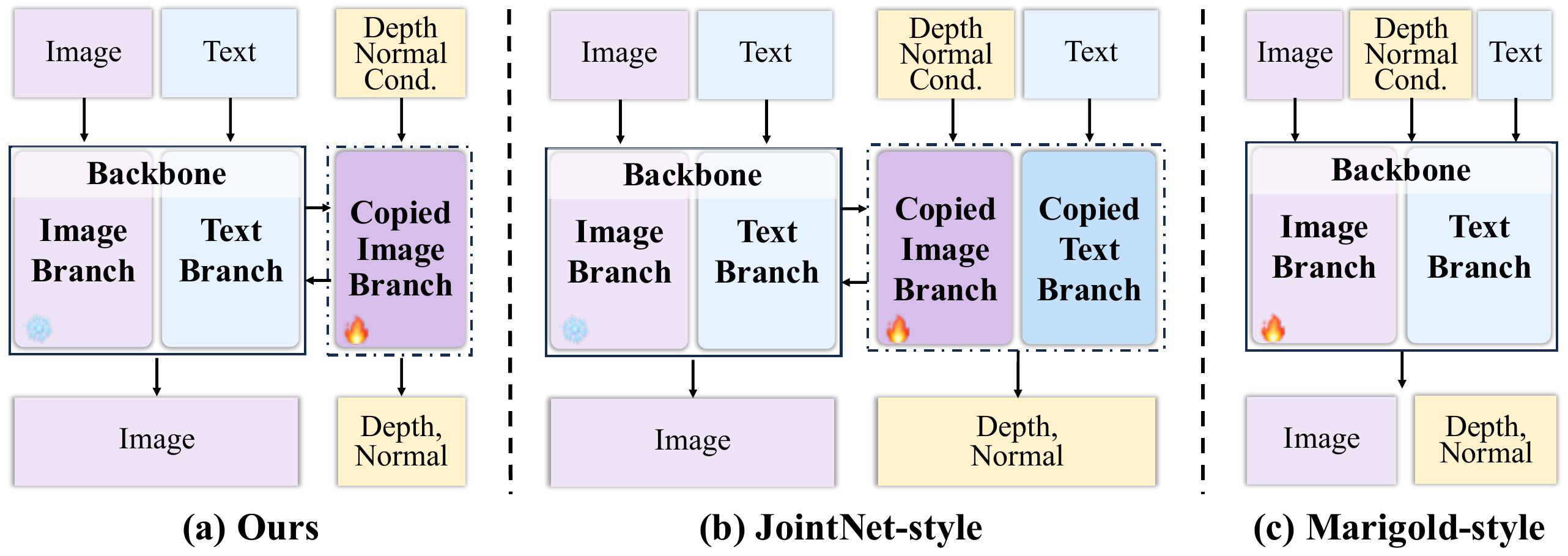}
  \caption{\textbf{Demonstration of representative design paradigms.}
    \modelname~copies only the image branch from MMDiT to model additional visual distributions while explicitly preserving the backbone’s versatile priors.
    JointNet-style duplicates the entire backbone, incurring heavy computation;
    Marigold-style fine-tunes the backbone itself, quickly forgetting generative priors.}
  \label{fig:design}
\end{figure*}

\subsection{Preliminaries and Problem Setting}
\label{Preliminary}
\noindent\textbf{Diffusion Transformer}~\cite{peebles2023scalable} replaces the commonly used U-Net backbone in diffusion models with Transformer~\cite{vaswani2017attention}.
DiT first converts spatial inputs into a sequence of tokens and then performs denoising through a series of transformer blocks.
Among these, MMDiT is a powerful variant widely adopted in recent SOTA visual generation models~\cite{sd3,Flux}.
Specifically, as shown in the dashed box of~\cref{fig:method:init}, MMDiT models text and image features in two separate branches, applying full attention only once in each transformer block. We denote the text and image branches as \( \model_t \) and \( \model_i \), respectively.

\noindent\textbf{Problem Setting.}
Our model is designed to jointly produce three outputs: an RGB image $\mathbf{x}$, depth map $\mathbf{d}$, and surface normal map $\mathbf{n}$, based on the input text prompt \(\bc_t\) and condition \(\bc\). We investigate three primary settings for this condition:
    \textit{\textbf{i) Controllable Generation.}} The condition $\mathbf{c}$ is a spatial map, such as \textit{Depth, Normal} or \textit{Sketch}. The model synthesizes $\mathbf{x}$ while simultaneously inferring its geometric properties ($\mathbf{d}$, $\mathbf{n}$).
    \textit{\textbf{ii) Perception.}} The condition $\mathbf{c}$ is an RGB image. Here, the model's goal is to predict its geometric pairs ($\mathbf{d}$, $\mathbf{n}$).
    \textit{\textbf{iii) Joint Generation.}} The condition consists solely of a text prompt, allowing for the joint generation of the image and its geometry purely from \(\bc_t\).
More formally, our objective is to learn a unified diffusion model \(\model(\bc_t, \bc) \).

\subsection{\textbf{\modelname:} Unified Generation and Perception}
\label{our_model}
\modelname~consists of two parts, with the overall framework illustrated in~\cref{fig:method:framework}. In~\cref{sec:model}, we introduce \methodname, which enables our framework to model joint distributions of RGB and its geometry pairs flexibly. In~\cref{sec:strategy}, we present the unified dataset and training strategy.

\subsubsection{Disentangled Unified Generation and Perception}
\label{sec:model}
To achieve unified generation and perception on the MMDiT architecture, one straightforward option is to duplicate the entire backbone, as in JointNet~\cite{zhang2024jointnet}. However, this doubles the parameters and incurs substantial computational overhead. Another option, following Marigold~\cite{ke2023repurposing}, is to directly fine-tune the backbone itself, but this quickly erases the versatile generative priors.

\begin{wrapfigure}{r}{0.47\textwidth}
    \vspace{-12pt}
    \centering
    \includegraphics[width=\linewidth]{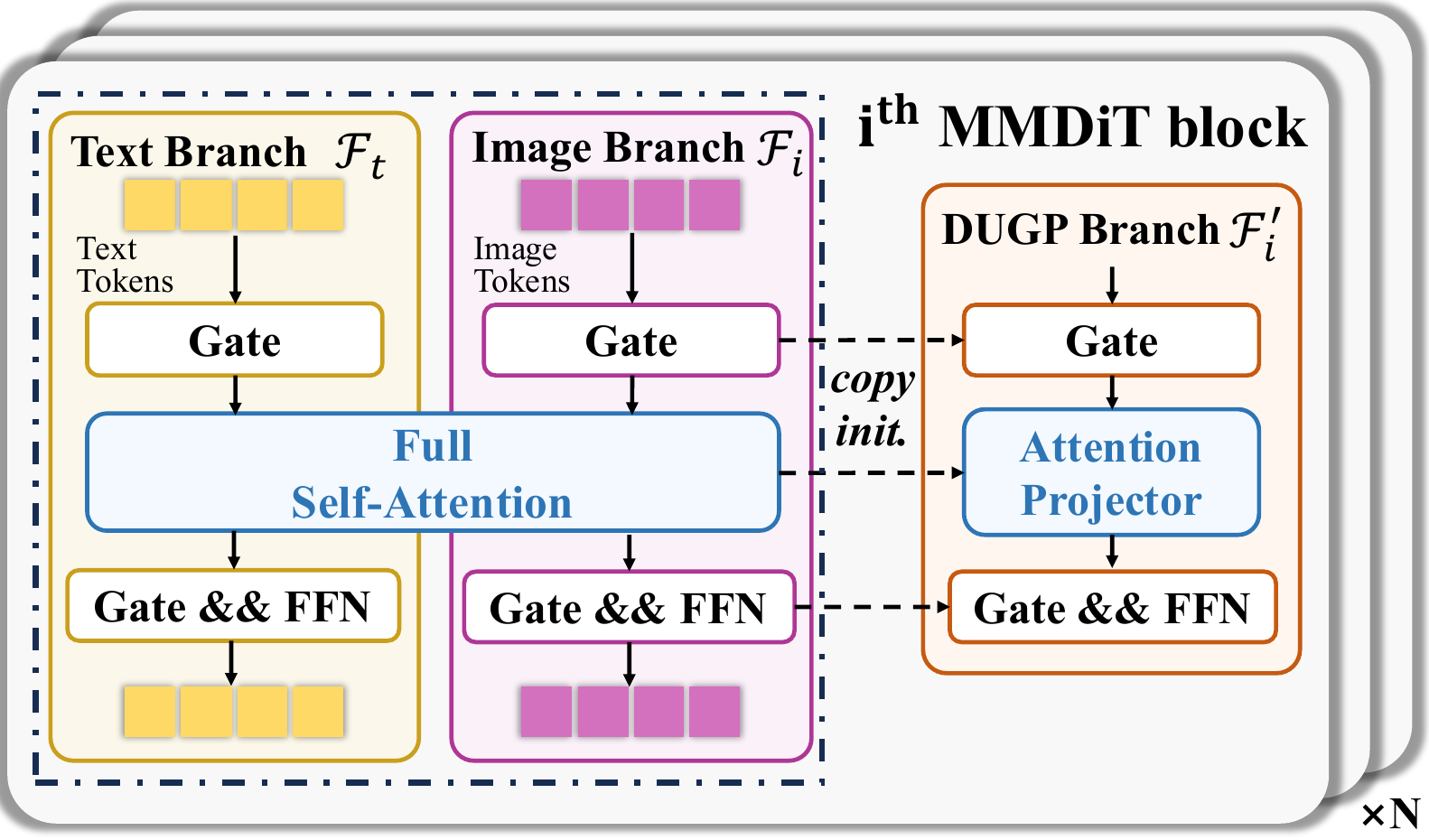}
    \captionsetup{font=small}
    \caption{
        \textbf{The MMDiT block and initialization of \methodname.} The {\textit{left dashed box}} represents the MMDiT, which consists of separate text and image branches.
        The {\textit{right dashed box}} represents \methodname~, which is initialized by copying the image branch of MMDiT.
    }
    \label{fig:method:init}
    \vspace{-14pt}
\end{wrapfigure}

We revisit the MMDiT architecture, as illustrated in the \textit{left dashed box} of~\cref{fig:method:init}.
MMDiT (denoted as \( \model \)) consists of a text branch \( \model_t \) and an image branch \( \model_i \). Although the text branch \( \model_t \) accounts for half the parameters, it cannot model the visual distribution.
Following the principle of Occam's razor, the image branch \( \model_i \) is the only necessary entirety to model additional visual distributions.
Thus, \textbf{we only copy a trainable image branch} \( \model'_i \) based on \( \model_i \) as an additional branch for perception-related visual distributions.
As shown in~\cref{fig:method:init} \textit{right}, this new branch is referred to as \methodname.
We divide \methodname~into a {condition patcher}, {stacked control layers}, and {stacked perception layers}, where the control and perception layers consist of \(m\) and \(n\) blocks respectively.

\noindent\textbf{Condition Patcher.}
To seamlessly integrate the condition \( \mathbf{c} \), we introduce a parallel processing path at the \methodname's input section.
First, we copy the patcher of the original model, $\model_i$. This new layer is zero-initialized to prevent the control signal from abruptly affecting the model during early training.
Concurrently, the original pre-trained patcher processes the noisy depth and normal latents.
Finally, the output tokens of noisy depth/normal latents and condition \( \mathbf{c} \) are concatenated along the sequence dimension and fed into \( \model'_i \).

\noindent\textbf{Stacked Control Layers.}
To control the backbone, \methodname~employs the attention mechanism and shares the backbone’s text branch \( \model_t \). The attention calculation between \methodname~and the backbone is as follows, where $\sigma(\cdot)$ denotes the Softmax function:
\begin{equation}
    [\mathbf{A}_d, \mathbf{A}_n] = \sigma\left(\frac{[\mathbf{Q}_d, \mathbf{Q}_n] [\mathbf{K}_t, \mathbf{K}_d, \mathbf{K}_n]^T}{\sqrt{d_k}}\right) [\mathbf{V}_t, \mathbf{V}_d, \mathbf{V}_n],
    \label{eq:attention:depth}
\end{equation}
\begin{equation}
    \mathbf{A}_c = \sigma\left(\frac{\mathbf{Q}_c [\mathbf{K}_t, \mathbf{K}_i, \mathbf{K}_c]^T}{\sqrt{d_k}}\right) [\mathbf{V}_t, \mathbf{V}_i, \mathbf{V}_c].
    \label{eq:attention:condition}
\end{equation}
In ~\cref{eq:attention:depth,eq:attention:condition}, $[\cdot, \cdot]$ denotes sequence concatenation.
At the end of each block in the stacked control layers, we obtain the output \( \mathbf{I} \) of the backbone's image branch \( \model_i \) and the condition's outputs of \( \model'_i \), denoted as \( \mathbf{C} \). We then add \( \mathbf{C} \) to \( \mathbf{I} \) through a zero-initialized linear layer $\mathcal{L}_{\text{zero}}$:
\begin{equation}
    \mathbf{I}' = \mathbf{I} + \mathcal{L}_{\text{zero}}(\mathbf{C}).    \label{eq:attention:add}
\end{equation}

\noindent\textbf{Stacked Perception Layers.}
In the stacked perception layers, our goal is to extract features from the input condition \( \mathbf{c} \) and the backbone to output the corresponding depth and normal. We modify the computation here as follows:
\begin{equation}
    [\mathbf{A}_d, \mathbf{A}_n] = \sigma\left( \frac{[\mathbf{Q}_d, \mathbf{Q}_n] [\mathbf{K}_d, \mathbf{K}_n, \mathbf{K}_t, \mathbf{K}_i, \mathbf{K}_c]^T}{\sqrt{d_k}} \right) [\mathbf{V}_d, \mathbf{V}_n, \mathbf{V}_t, \mathbf{V}_i, \mathbf{V}_c].
    \label{eq:attention:perception}
\end{equation}
Notably, the feature addition from~\cref{eq:attention:add} is bypassed in the perception layers. This architectural choice forms the core mechanism for our ``Prior Preserved'' objective via strict gradient isolation.
The fundamental dilemma in unifying generation and dense prediction lies in their conflicting optimization objectives: perception requires deterministic, discriminative pixel-level mapping, which often collapses the highly diverse distributions learned by generative models.
By severing the additive $\mathcal{L}_{\text{zero}}$ path for perception layers, gradients originating from the dense prediction loss flow exclusively through the \methodname~branch $\model'_i$. This prevents them from altering the pre-trained weights of the backbone $\model_i$. The backbone acts purely as a robust feature donor, forcing the perception layers to learn as non-invasive ``readers'' and ensuring that multi-modal generative priors are perfectly quarantined and preserved.

\noindent\textbf{Handling Multiple Input Conditions.}
For scenarios requiring multiple simultaneous conditions, we duplicate the \methodname~condition patcher for each condition. Crucially, the interaction with the backbone is performed via sequence concatenation along the same dimension. This means all condition latents are processed within a \textit{single forward pass} of the backbone, effectively preventing the inference time from scaling linearly with the number of conditions and maintaining high efficiency.

\clearpage
\subsubsection{Unified Dataset Training Strategy}
\label{sec:strategy}
Diffusion generation and perception tasks traditionally use separate datasets. To unify these domains, our dataset integrates both, combining filtered generation data~\cite{qin2023unicontrol} with filtered synthetic perception data~\cite{roberts2021hypersim,cabon2020virtual}.
We address the heterogeneous nature of this data by randomly constructing batches on the fly and adjusting the loss weight adaptively based on the dataset origin.

Specifically, we learn a network \( v_\theta \) that predicts the velocity field~\cite{sd3,lipman2022flow} \( u_t \) given the image condition \( \mathbf{c} \) and text prompt \( \mathbf{c}_t \). We minimize the training objective as follows:
\begin{equation}
\label{eq:loss}
\begin{split}
    \mathcal{L} &= \mathbb{E}_{t, x, d, n, \mathbf{c}_t, \mathbf{c}} \Big[ \lambda_g \| v_\theta(x_t, t, \mathbf{c}_t, \mathbf{c}) - u_t( x_t, t \mid x_1 ) \|^2 \\
    &\hspace{-0.58em} + \lambda_p \| v_\theta(d_t, t, \mathbf{c}_t, \mathbf{c}) - u_t( d_t, t \mid d_1 ) \|^2 + \lambda_p \| v_\theta(n_t, t, \mathbf{c}_t, \mathbf{c}) - u_t( n_t, t \mid n_1 ) \|^2 \Big].
\end{split}
\end{equation}
Here, \(x_1\), \(d_1\), and \(n_1\) represent the latents for RGB, depth, and normal, respectively. The routing weights \(\lambda_g, \lambda_p \in \{0, 1\}\) act as binary task selectors. Rather than manually balancing complex task-specific loss scales, we dynamically assign these weights based on the data source of each sample within the mixed batch. This straightforward yet highly effective strategy allows us to jointly optimize tasks on a strictly heterogeneous dataset. The complete joint training pipeline, detailing this adaptive sample-level routing, is summarized in \cref{alg:training}.

\begin{algorithm}[H]
\caption{Joint Training Strategy of \modelname}
\label{alg:training}
\begin{algorithmic}[1]
\REQUIRE Pre-trained MMDiT backbone $\model$, initialized \methodname~branch $\model'$, Generation Dataset $\mathcal{D}_g$, Perception Dataset $\mathcal{D}_p$
\ENSURE Optimized \methodname~parameters $\theta$
\WHILE{not converged}
    \STATE Sample a mixed batch $B = B_g \cup B_p$ where $B_g \sim \mathcal{D}_g$ and $B_p \sim \mathcal{D}_p$
    \FOR{each sample $s_i \in B$}
        \IF{$s_i \in B_g$}
            \STATE Set loss weights: $\lambda_g^{(i)} \leftarrow 1$, $\lambda_p^{(i)} \leftarrow 0$
            \STATE Set condition $\mathbf{c} \leftarrow$ random spatial condition or NONE
        \ELSE
            \STATE Set loss weights: $\lambda_g^{(i)} \leftarrow 0$, $\lambda_p^{(i)} \leftarrow 1$
            \STATE Set condition $\mathbf{c} \leftarrow$ Original RGB image
        \ENDIF
        \STATE Sample timestep $t \sim \mathcal{U}(0, 1)$
        \STATE Add noise to latents $(x_1, d_1, n_1)$ to get $(x_t, d_t, n_t)$
        \STATE Forward pass backbone $\model$ and \methodname~$\model'$ with $\mathbf{c}_t$ and $\mathbf{c}$
        \STATE Compute flow matching loss $\mathcal{L}^{(i)}$ using Eq. (\ref{eq:loss})
    \ENDFOR
    \STATE Compute batch gradient $\nabla_\theta \frac{1}{|B|} \sum_i \mathcal{L}^{(i)}$
    \STATE Update parameters $\theta$ of \methodname~using Adam optimizer
\ENDWHILE
\end{algorithmic}
\end{algorithm}

\begin{figure}[t]
  \centering
  \includegraphics[width=\textwidth]{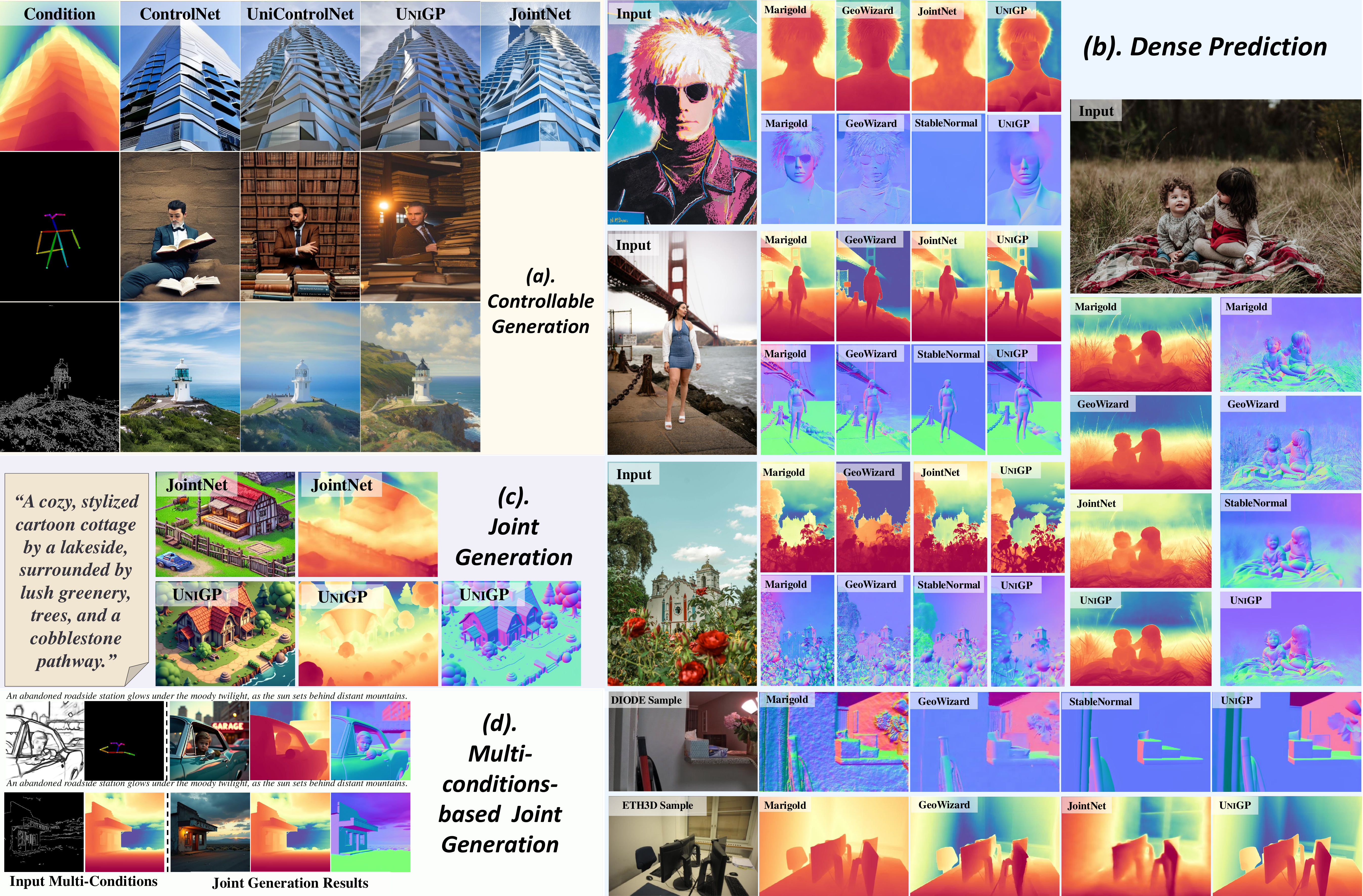}
  \caption{\textbf{Qualitative comparison and results} on \textbf{(a)} controllable generation, \textbf{(b)} dense prediction, \textbf{(c)} joint-generation and \textbf{(d)} multi-condition-based joint generation tasks between \modelname~and representative diffusion-based methods.
  \modelname~outperforms previous diffusion-based experts and unified models across all tasks.
  }
  \label{fig:method:qualitative}
  \vspace{-10pt}
\end{figure}

\section{Experiments}
\label{sec:experiments}

\subsection{Experimental Settings}
\label{sec:setting}

\noindent\textbf{Implementation Details.}
We implement \modelname~on SD3-medium and optimize using the Adam optimizer with a learning rate of $1 \times 10^{-4}$. The backbone is frozen, meaning we only train the \methodname~branch. All experiments are conducted on 16 NVIDIA A800 GPUs with a total batch size of 64, for 40K steps.

\noindent\textbf{Training Datasets.}
Our unified dataset integrates two heterogeneous data sources.
\textbf{1) Perception Component:} We use synthetic datasets covering both indoor and outdoor scenes to ensure perfect geometric ground truth. Specifically, we use 39K complete samples filtered from Hypersim (indoor) resized to $576 \times 768$, and 20K samples from Virtual KITTI 2 (outdoor) with the far plane set at 80m.
\textbf{2) Generation Component:} We filter and randomly select 1M samples from MultiGen-20M, ensuring a minimum edge length of 768. We utilize its existing annotations (\textit{normal, canny, sketch, human pose}). Importantly, we re-annotate the \textit{depth} using GeoWizard because original discriminative annotations like MiDaS or DepthAnything output inverse normalized depth, which is structurally incompatible with diffusion-based generation.
During training, perception samples use the original RGB image as the condition, while generation samples randomly select a spatial condition or NONE (for joint-generation) with equal probability.

\noindent\textbf{Evaluation Datasets and Metrics.}
\textbf{1) Perception Tasks:} For zero-shot affine-invariant depth estimation, we evaluate on NYUv2, ScanNet, KITTI, and ETH3D using absolute mean relative error (AbsRel) and inlier metrics ($\delta_1, \delta_2$). For surface normal estimation, we test on NYUv2, ScanNet, iBims-1, and Sintel, reporting mean angular error (m.) and accuracy ($11.25^\circ, 30^\circ$).
\textbf{2) Generation Tasks:} We evaluate on 5K samples from the MSCOCO validation set using Fréchet Inception Distance (FID) and CLIP Score.
\textbf{3) Cross-modal Consistency Metrics:} Existing metrics like FID or CLIP score primarily evaluate the visual quality or text-alignment of generated images, completely ignoring whether the jointly generated RGB and geometries are structurally aligned. To explicitly measure this geometric consistency in joint-generation tasks, we propose \textbf{GD} (\textbf{G}enerated \textbf{D}epth RMSE) and \textbf{GN} (\textbf{G}enerated \textbf{N}ormal RMSE). These are computed by passing the generated RGB through an expert model (GeoWizard) to recalculate the pseudo-ground-truth depth/normal, and then calculating the RMSE against our model's directly generated depth/normal. Lower GD/GN strictly indicates higher structural alignment and geometric consistency. For controllable generation, RMSE is calculated directly between the generated geometry and the input condition.

\noindent\textbf{Comparison Methods.}
We benchmark \modelname~against state-of-the-art methods across multiple domains. \textbf{1) Controllable Generation:} We compare with single-control models including ControlNet~\cite{zhang2023adding} and SD3-ControlNet~\cite{sd3_controlnet} (\ie SD3-CNet), as well as multi-control models including UniControlNet~\cite{zhao2024uni} and ControlNetPlus~\cite{ControlNetPlus} (\ie CNetPlus). \textbf{2) Joint Generation:} We evaluate against recent unified frameworks, including LDM3D~\cite{stan2023ldm3d}, JointNet~\cite{zhang2024jointnet}, and UniCon~\cite{li2024unicon}. \textbf{3) Dense Prediction:} For depth and surface normal estimation, we comprehensively compare against standard regression experts (MiDaS~\cite{ranftl2020towards}, Omnidata V1/V2~\cite{eftekhar2021omnidata,kar20223d}, DPT~\cite{ranftl2021vision}, DepthAnything V1/V2~\cite{yang2024depth1,yang2024depth2}, and DSINE~\cite{bae2024rethinking}) as well as recent generative perception methods (Marigold~\cite{ke2023repurposing}, GeoWizard~\cite{fu2025geowizard}, StableNormal~\cite{ye2024stablenormal}, OneDiffusion~\cite{le2025one}, and JoDi~\cite{xu2025jodi}).

\subsection{Main Results}
\label{sec:results}

\subsubsection{Qualitative evaluation}
\label{sec:qualitative}

In~\cref{fig:method:qualitative}, we show main results generated by \modelname~on different tasks and compare them with representative methods.
\textit{Note that all our results are generated by a single model.}
\textbf{1)} First, in controllable generation, our method supports more conditions than ControlNet and achieves better generation quality and text-image consistency than UniControlNet.
\textbf{2)} Second, in perception tasks, \modelname~outperforms previous generative-based dense prediction methods, providing fine-grained details and accurate depth/normal estimations for complex geometries. Compared to the unified model JointNet, \modelname~simultaneously estimates both depth and normals, achieving significantly higher depth estimation accuracy.
\textbf{3)} Third, in terms of joint-generation, JointNet can only generate RGB and depth simultaneously, whereas \modelname~is capable of generating RGB, depth, and normal comprehensively.

\subsubsection{Quantitative evaluation}

We comprehensively evaluate the quantitative performance of \modelname~across generation and perception tasks.

\noindent\textbf{Generation Results.}
We show quantitative joint generation results in~\cref{tab:generation_joint} and a comprehensive evaluation across diverse controllable generation tasks in~\cref{tab:generation_control}.
\textbf{1)} In terms of joint-generation, \modelname~outperforms the baseline models across all metrics by a large margin, demonstrating our method's superiority in preserving the backbone's generative capabilities while maintaining high geometry consistency.
\textbf{2)} In controllable generation, \modelname~consistently achieves the leading FID and CLIP scores across dense geometric modalities (\eg depth, normal) and sparse structural modalities (\eg sketch, pose). Notably, it achieves the lowest RMSE between generated images and the given conditions in depth-to-image and normal-to-image tasks, demonstrating rigorous spatial and structural alignment.

\begin{table}[H]
    \centering
    \caption{\textbf{Comparison on joint generation.}
    \textbf{Bold} and \underline{underline} denote the best and second-best results, respectively.}
    \label{tab:generation_joint}
    \resizebox{0.62\linewidth}{!}{
    \begin{tabular}{lccccc}
        \toprule
        \multirow{2}{*}{\textbf{Methods}} &
        \textbf{Training} &
        \multicolumn{4}{c}{\textbf{Joint Generation Results}} \\
        \cmidrule(lr){3-6}
        &\textbf{Data} & FID $\downarrow$ & CLIP $\uparrow$ & GD $\downarrow$ & GN $\downarrow$ \\
        \midrule
        LDM3D & - & 30.19 & 26.02 & 18.13 & - \\
        JointNet & 2.56M & \underline{28.02} & \underline{27.00} & \underline{16.80} & - \\

        \textbf{\modelname~(Ours)} & 1M & \textbf{21.05} & \textbf{31.50} & \textbf{6.05} & \textbf{6.41} \\
        \bottomrule
    \end{tabular}
    }
\end{table}

\noindent\textbf{Perception Results.}
\textbf{1)} We present depth estimation results in~\cref{tab:depth}, where \modelname~achieves the best performance among all generative baselines and is on par with the SOTA regression DepthAnything series on indoor/outdoor datasets, and outperforms them on diverse datasets (\eg ETH3D).
This highlights the power of generative priors. Unlike regression methods trained discriminatively from scratch on 62.6M images, \modelname~trains perception on only 0.059M images. By retaining the rich world knowledge of the SD3 backbone, it achieves superior zero-shot generalization. Notably, JointNet's performance lags far behind, ranking 10th.
\textbf{2)} Surface normal estimation results in~\cref{tab:normal} show that \modelname~achieves comparable performance to DSINE, a recent SOTA regression model, and surpasses all generative methods.

\noindent\textbf{Generalization to Diverse Tasks.}
Beyond the core tasks evaluated above, a fundamental advantage of embedding perception into a generative framework is the inherent versatility of the generative latent space. As detailed in the supplementary material, \modelname~can be easily extended to a broader range of perception tasks, such as semantic segmentation, albedo estimation, and shading estimation. This demonstrates that our \methodname~branch effectively captures a generalized representation of physical world properties.

\begin{table}[ht!]
\caption{
\textbf{Comparison of \modelname~with task-specific baselines on controllable generation tasks.}
We evaluate across five spatial conditions. $^{*}$ denotes single-control methods, $^{\dagger}$ denotes multi-control methods, and $^{\ddagger}$ denotes joint-generation methods.
}
\label{tab:generation_control}
\centering
\resizebox{\textwidth}{!}{
\begin{tabular}{ll ccccccc}
    \toprule
    \textbf{Condition} & \textbf{Metric} & \textbf{ControlNet$^{*}$} & \textbf{SD3-CNet$^{*}$} & \textbf{UniControlNet$^{\dagger}$} & \textbf{CNetPlus$^{\dagger}$} & \textbf{JointNet$^{\ddagger}$} & \textbf{UniCon$^{\ddagger}$} & \textbf{\modelname~(Ours)$^{\ddagger}$} \\
    \midrule
    \multicolumn{2}{l}{\textit{Training Data}} & - & - & 10M & 10M & 2.56M & 16K & 1M \\
    \midrule
    \multirow{3}{*}{\textbf{Depth}}
    & FID $\downarrow$  & 19.80 & \underline{18.00} & 20.09 & 19.27 & 25.66 & 19.21 & \textbf{17.95} \\
    & CLIP $\uparrow$ & 25.30 & 27.09 & 25.25 & \underline{27.99} & 25.09 & 25.27 & \textbf{28.41} \\
    & RMSE $\downarrow$ & 13.86 & \underline{12.50} & 15.93 & 13.20 & 14.88 & 12.91 & \textbf{6.89} \\
    \midrule
    \multirow{3}{*}{\textbf{Normal}}
    & FID $\downarrow$  & 22.18 & - & - & \underline{20.11} & - & 21.35 & \textbf{18.60} \\
    & CLIP $\uparrow$ & 25.10 & - & - & \underline{26.80} & - & 25.12 & \textbf{28.45} \\
    & RMSE $\downarrow$ & 18.15 & - & - & \underline{16.00} & - & 16.55 & \textbf{7.12} \\
    \midrule
    \multirow{2}{*}{\textbf{Canny}}
    & FID $\downarrow$  & 16.16 & 17.92 & 17.79 & \underline{16.12} & - & 17.84 & \textbf{15.91} \\
    & CLIP $\uparrow$ & 25.34 & 26.88 & 25.39 & \underline{27.90} & - & 25.31 & \textbf{30.35} \\
    \midrule
    \multirow{2}{*}{\textbf{Sketch}}
    & FID $\downarrow$  & \underline{17.35} & 19.12 & 18.55 & 17.44 & - & 18.62 & \textbf{17.02} \\
    & CLIP $\uparrow$ & 25.45 & 27.20 & 25.52 & \underline{28.15} & - & 25.18 & \textbf{30.01} \\
    \midrule
    \multirow{2}{*}{\textbf{Pose}}
    & FID $\downarrow$  & 23.42 & 20.18 & 22.90 & \underline{18.95} & - & 21.55 & \textbf{15.40} \\
    & CLIP $\uparrow$ & 25.25 & 27.45 & 25.38 & \underline{28.60} & - & 25.22 & \textbf{30.79} \\
    \bottomrule
\end{tabular}
}
\vspace{-15pt}
\end{table}

\begin{table}[ht!]
\caption{\textbf{Quantitative comparison on zero-shot affine-invariant depth estimation.} We compare \modelname~with both regression and generative SOTA methods.
\underline{Underline} marks the overall best, and \textbf{Bold} marks the best among generative methods.
Note: Generative baselines utilize different backbones.
}
\label{tab:depth}
\centering
\resizebox{\textwidth}{!}{
\begin{tabular}{lccccccccccc}
\toprule
\multirow{2}{*}{Method} & Training
& \multicolumn{2}{c}{NYUv2 (Indoor)} & \multicolumn{2}{c}{KITTI (Outdoor)}
& \multicolumn{2}{c}{ETH3D (Various)} & \multicolumn{2}{c}{ScanNet (Indoor)}
& Avg.\\
\cmidrule(lr){3-4} \cmidrule(lr){5-6} \cmidrule(lr){7-8} \cmidrule(lr){9-10}
 & Data
 & AbsRel$\downarrow$ & $\delta$1$\uparrow$ & AbsRel$\downarrow$ & $\delta$1$\uparrow$
 & AbsRel$\downarrow$ & $\delta$1$\uparrow$ & AbsRel$\downarrow$ & $\delta$1$\uparrow$
 & Rank   \\
\midrule
\rowcolor{mylightgray} \multicolumn{11}{l}{\textbf{Regression Methods}} \\
\midrule
MiDaS
 &2M
& 11.1 & 88.5 & 23.6 & 63.0
& 18.4 & 75.2 & 12.1 & 84.6
&  9.88        \\
Omnidata
 &12.2M
& 7.4 & 94.5 & 14.9 & 83.5
& 16.6 & 77.8 & 7.5 & 93.6
& 6.50 \\

DPT
 &1.4M
& 9.8 & 90.3 & 10.0 & 90.1
& 7.8 & 94.6 & 8.2 & 93.4
&  6.50        \\

DepthAnything
&62.6M
& \underline{4.3} & \underline{98.1}
& 7.6 & \underline{94.7}
& 12.7 & 88.2
& 4.3 & \underline{98.1}
& \underline{2.25}  \\

DepthAnything V2
 &62.6M
 & 4.5 & 97.9
 & \underline{7.4} & 94.6
 & 13.1 & 86.5
 & \underline{4.2} & 97.8
 &  2.88   \\

\midrule
\rowcolor{mylightgray} \multicolumn{11}{l}{\textbf{Generative Methods}} \\
\midrule

GeoWizard
 &280K
& 5.6 & 96.3 & 14.4 & 82.0
& 6.6 & 95.8 & 6.4 & 95.0
&  4.81         \\

Marigold
 &74K
& 5.5 & 96.4 & 9.9 & 91.6
& 6.5 & 95.9 & 6.4 & 95.2
&    3.56         \\

JointNet
 &2.56M
& 13.6 & 84.1 & 29.9 & 59.6
& 19.2 & 78.7 & 11.9 & 84.8
&    10.00         \\

UniCon
 &16K
& 7.9 & 93.9 & - & -
& - & - & 9.2 & 91.9
&    7.50         \\

OneDiffusion
 &75M
& 8.9 & 92.0 & - & -
& - & - & 9.7 & 90.7
&    8.88         \\

JoDi
 &290K
& 8.3  & 92.0 & - & -
& - & - & 9.9 & 90.3
&    9.13         \\

\textbf{\modelname~ (Ours)} & 59K
& \textbf{5.2} & \textbf{96.6} & \textbf{8.3} & \textbf{93.3}
& \textbf{\underline{6.0}} & \textbf{\underline{96.3}} & \textbf{5.5} & \textbf{97.9}
& \textbf{2.38} \\

\bottomrule
\end{tabular}
}
\end{table}

\begin{table}[t]
\caption{\textbf{Quantitative comparison on zero-shot surface normal estimation.}
Note: Generative baselines utilize different backbones.
}
\label{tab:normal}
\centering
\resizebox{\textwidth}{!}{
\begin{tabular}{lccccccccccccccc}
\toprule
\multirow{2}{*}{Method} & Training
& \multicolumn{3}{c}{NYUv2 (Indoor)} & \multicolumn{3}{c}{ScanNet (Indoor)}
& \multicolumn{3}{c}{iBims-1 (Indoor)} & \multicolumn{3}{c}{Sintel (Outdoor)}
& Avg. \\
\cmidrule(lr){3-5} \cmidrule(lr){6-8} \cmidrule(lr){9-11} \cmidrule(lr){12-14}
 &Data
 & m.$\downarrow$ & $11.25^\circ$$\uparrow$ & $30^\circ$$\uparrow$
 & m.$\downarrow$ & $11.25^\circ$$\uparrow$ & $30^\circ$$\uparrow$
 & m.$\downarrow$ & $11.25^\circ$$\uparrow$ & $30^\circ$$\uparrow$
 & m.$\downarrow$ & $11.25^\circ$$\uparrow$ & $30^\circ$$\uparrow$
 & Rank         \\
\midrule
\rowcolor{mylightgray} \multicolumn{15}{l}{\textbf{Regression Methods}} \\
\midrule
Omnidata
 &12.2M
& 23.1 & 45.8 & 73.6 & 22.9 & 47.4 & 73.2 & 19.0 & 62.1 & 80.1 & 41.5 & 11.4 & 42.0
&  6.92         \\
Omnidata V2
 &12.2M
& 17.2 & 55.5 & 83.0 & 16.2 & 60.2 & 84.7 & 18.2 & 63.9 & 81.1 & 40.5 & 14.7 & 43.5
&  3.50         \\
DSINE
&160K
& \underline{16.4} & \underline{59.6} & \underline{83.5}  & 16.2 & 61.0 & 84.4 & \underline{17.1} & \underline{67.4} & 82.3 & \underline{34.9} & \underline{21.5} & 52.7
& 1.67 \\
\midrule
\rowcolor{mylightgray} \multicolumn{15}{l}{\textbf{Generative Methods}} \\
\midrule
Marigold
&74K
& 20.9 & 50.5 & - & 21.3 & 45.6 & - & 18.5 & 64.7 & - & -    & -    & -
& 7.40          \\
GeoWizard
&280K
& 18.9 & 50.7 & 81.5 & 17.4 & 53.8 & 83.5 & 19.3 & 63.0 & 80.3 & 40.3 & 12.3 & 43.5
& 5.29          \\
StableNormal
&250K
& 18.6 & 53.5 & 81.7 & 17.1 & 57.4 & 84.1 & 18.2 & 65.0 & 82.4 & 36.7 & 14.1 & 50.7
& 3.63 \\
JoDi
&290K
& 18.6 & - & - & 20.3 & - & - & 18.2 & - & - & - & - & -
& 4.83 \\
\textbf{\modelname~ (Ours)} & 59K
& \textbf{\underline{16.4}} & \textbf{59.2} & \textbf{83.4}
& \textbf{\underline{14.9}} & \textbf{\underline{65.1}} & \textbf{\underline{86.0}}
& \textbf{17.3} & \textbf{66.5} & \textbf{\underline{82.8}}
& \textbf{35.0} & \textbf{20.1} & \textbf{\underline{55.1}}
& \textbf{\underline{1.58}} \\
\bottomrule
\end{tabular}
}
\end{table}

\begin{figure}[t]
  \centering
  \includegraphics[width=\textwidth]{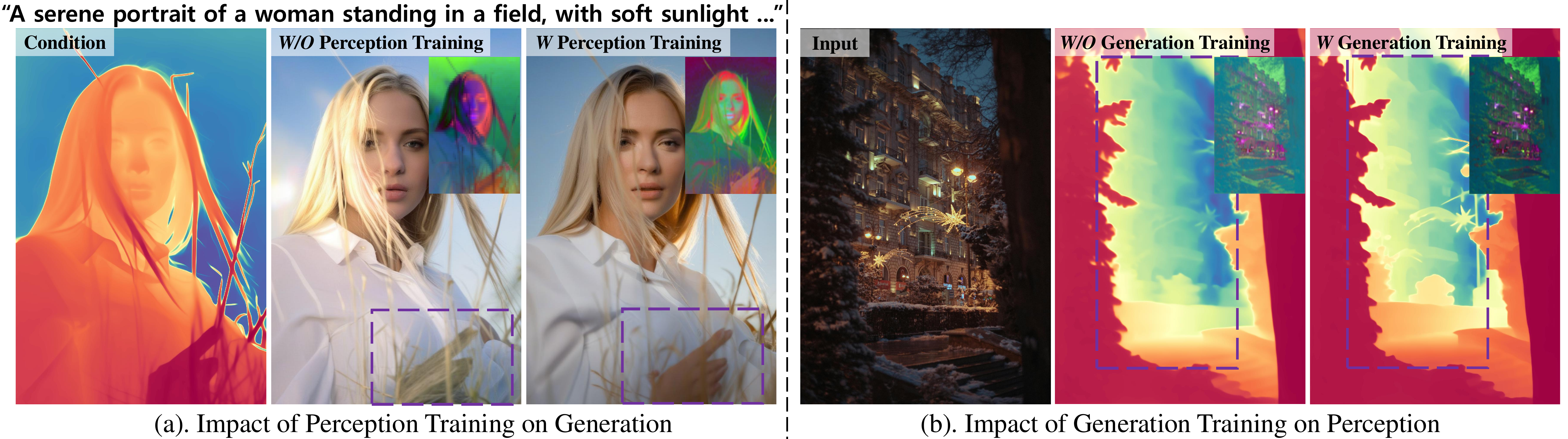}
  \caption{\textbf{Ablation study on the relationship between generation and perception.} Comparison areas are highlighted with \textcolor{customPurple}{purple boxes}. Adding perception training makes the generation results strictly align with the condition boundaries. Adding generation training improves the perception results with finer details.}
  \label{fig:ablation}
\vspace{-15pt}
\end{figure}

\begin{table}[ht!]
\caption{\textbf{Impact of Joint Training and Dataset Strategy.} Evaluating the complementary benefits of cross-task training on representative benchmarks.}
\label{tab:ablation_relation}
\centering
\resizebox{\textwidth}{!}{
    \begin{tabular}{lcccccccccccccc}
        \toprule
        \multirow{2}{*}{\textbf{Methods}} &
        \multicolumn{4}{c}{\textbf{Text to Image}} &
        \multicolumn{3}{c}{\textbf{Depth to Image}} &
        \multicolumn{3}{c}{\textbf{Normal to Image}} &
        \multicolumn{2}{c}{\textbf{Depth Esti.}} &
        \multicolumn{2}{c}{\textbf{Normal Esti.}} \\
        \cmidrule(lr){2-5} \cmidrule(lr){6-8} \cmidrule(lr){9-11} \cmidrule(lr){12-13} \cmidrule(lr){14-15}
        & FID $\downarrow$ & CLIP $\uparrow$ & GD $\downarrow$ & GN $\downarrow$ & FID $\downarrow$ & CLIP $\uparrow$ & RMSE $\downarrow$ & FID $\downarrow$ & CLIP $\uparrow$ & RMSE $\downarrow$ & AbsRel$\downarrow$ & $\delta$1$\uparrow$ & m.$\downarrow$ & $11.25^\circ$$\uparrow$  \\
    \midrule
        \textit{w/o} Perception Train. & 21.07 & 31.57 & - & - & 17.99 & 28.33 & 10.73 & 18.85  & 28.19  & 11.74 & - & - & - & - \\
        $50\%$ Perception Data  & 21.33 & 31.46 & 6.91 & 7.98 & 18.10 & 28.17 & 7.86 & 18.93  & 28.25  & 9.37 & 5.9 & 95.4 & 19.6 & 54.3 \\
        \textit{w/o} Generation Train. & - & - & - & - & - & - & - & -  & -  & - & 5.5 & 96.5 & 17.1 & 56.0 \\
        $50\%$ Generation Data  & 21.86 & 31.40 & 6.97 & 8.06 & 18.97 & 28.12 & 8.40 & 19.91 & 28.14  & 9.72 & 5.9 & 95.1 & 19.7 & 55.6 \\
        \textit{w/o} Training Strategy & 25.59 & 29.29 & 11.14 & 13.05 & 23.21 & 26.61 & 8.74 & 24.45  & 27.14  & 10.36 & 6.9 & 94.3 & 22.6 & 46.9 \\
        \textbf{\modelname~(Ours)} & \textbf{21.05} & \textbf{31.50} & \textbf{6.05} & \textbf{6.41} & \textbf{17.95} & \textbf{28.41} & \textbf{6.89} & \textbf{18.60} & \textbf{28.45} & \textbf{7.12} & \textbf{5.2} & \textbf{96.6} & \textbf{16.4} & \textbf{59.2} \\
    \bottomrule
    \end{tabular}}
\vspace{-15pt}

\end{table}

\subsection{Ablation Study}
\label{sec:ablation}

In this section, we conduct ablation studies to evaluate \modelname. Due to computational constraints, the ablations are performed on representative benchmarks: generation on COCO-5K and perception on NYUv2.

\noindent\textbf{Relation between Generation and Perception.}
In~\cref{tab:ablation_relation}, we analyze the impact of isolating perception from generation, reducing the training data scale, and omitting our task-specific training strategy. The results demonstrate clear and strong \textbf{complementary benefits} between the two domains.

\textbf{1) Quantitative Improvements:} Incorporating perception training vastly improves the structural alignment of controllable generation, evidenced by the steep drop in depth-to-image RMSE from 10.73 down to 6.89. Conversely, incorporating massive generation training data imbues the perception layers with richer generative priors, pushing the limits of zero-shot perception metrics such as AbsRel and mean angular error by capturing finer and generalized details. Furthermore, omitting our task-specific batching strategy or halving the data scale leads to noticeable degradation across both domains, validating the necessity of our unified dataset construction.

\textbf{2) Deep Dive into Cross-Task Synergy:} To intuitively understand these quantitative gains, we provide a qualitative analysis in~\cref{fig:ablation}. When the model is trained exclusively on generation tasks without perception training, the generated pixels often bleed across the conditional spatial constraints. As highlighted by the purple boxes in~\cref{fig:ablation}(a), the model fails to strictly distinguish the foreground hand from the surrounding grass defined in the depth condition. However, once perception training is introduced, the model develops explicit spatial consciousness, forcing the generation results to strictly align with the condition boundaries. Conversely, as shown in~\cref{fig:ablation}(b), pure perception training tends to produce overly smoothed depth maps. The addition of generation training injects robust and high-frequency texture priors into the perception branch, enabling the model to capture realistic micro-structures, such as the intricate architectural details of the building facade, that are otherwise completely lost in standard regression models.

\begin{table}[ht!]
\vspace{-6pt}
\caption{\textbf{Balance between Stacked Control and Perception Layers.} We evaluate different block allocation ratios \textit{\((m,n)\)} where \(m+n=24\).}
\label{tab:ablation_balance}
\centering
\resizebox{\textwidth}{!}{
    \begin{tabular}{lcccccccccccccc}
        \toprule
        \multirow{2}{*}{\textbf{Methods}} &
        \multicolumn{4}{c}{\textbf{Text to Image}} &
        \multicolumn{3}{c}{\textbf{Depth to Image}} &
        \multicolumn{3}{c}{\textbf{Normal to Image}} &
        \multicolumn{2}{c}{\textbf{Depth Esti.}} &
        \multicolumn{2}{c}{\textbf{Normal Esti.}} \\
        \cmidrule(lr){2-5} \cmidrule(lr){6-8} \cmidrule(lr){9-11} \cmidrule(lr){12-13} \cmidrule(lr){14-15}
        & FID $\downarrow$ & CLIP $\uparrow$ & GD $\downarrow$ & GN $\downarrow$ & FID $\downarrow$ & CLIP $\uparrow$ & RMSE $\downarrow$ & FID $\downarrow$ & CLIP $\uparrow$ & RMSE $\downarrow$ & AbsRel$\downarrow$ & $\delta$1$\uparrow$ & m.$\downarrow$ & $11.25^\circ$$\uparrow$  \\
    \midrule
        \textit{\((m,n)\) = (0,24)} & - & - & - & - & - & - & - & -  & -  & - & 5.4 & 96.3 & 17.2 & 58.0 \\
        \textit{\((m,n)\) = (6,18)} & 25.11 & 29.85 & 9.15 & 9.87 & 21.15 & 28.04 & 9.21 & 23.25  & 27.41  & 10.25 & 5.3 & 96.0 & 17.1 & 57.6 \\
        \textit{\((m,n)\) = (18,6)} & 24.83 & 29.48 & 10.38 & 11.43 & 20.91 & 28.33 & 7.85 & 20.09  & 28.48  & 9.28 & 5.9 & 95.4 & 18.2 & 54.9 \\
        \textit{\((m,n)\) = (24,0)} & 21.78 & 31.37 & 7.56 & 7.95 & 19.63 & 28.18 & 10.65 & 18.91  & 28.21  & 10.49 & - & - & - & - \\
        \textbf{\modelname~(12,12)} & \textbf{21.05} & \textbf{31.50} & \textbf{6.05} & \textbf{6.41} & \textbf{17.95} & \textbf{28.41} & \textbf{6.89} & \textbf{18.60} & \textbf{28.45} & \textbf{7.12} & \textbf{5.2} & \textbf{96.6} & \textbf{16.4} & \textbf{59.2} \\
    \bottomrule
    \end{tabular}}
\vspace{-6pt}
\end{table}

\noindent\textbf{Balance between Stacked Control and Perception Layers.}
We denote the number of stacked control/perception layers as \((m,n)\), with the default setting \(m\)=\(n\)=12 for the 24-layer SD3 backbone. As shown in~\cref{tab:ablation_balance}, the optimal joint performance is consistently achieved when the capacities are balanced. Specifically, allocating too few control layers (\eg, \(m=6\)) severely degrades the controllable generation quality, as the network lacks the required depth to effectively inject and process spatial conditions. Conversely, allocating too few perception layers (\eg, \(n=6\)) significantly harms the dense prediction accuracy, indicating that extracting fine-grained geometric representations demands sufficient feature transformations. This insight implies that an equal capacity allocation allows the modules to co-develop a robust shared representation space, thereby maximizing cross-task synergy.

\begin{table}[ht!]
\vspace{-6pt}
\caption{\textbf{Model Design Choices.} Comparing \methodname~against alternative unified network paradigms, all implemented on the exact same SD3-Medium backbone.}
\label{tab:ablation_design}
\centering
\resizebox{\textwidth}{!}{
    \begin{tabular}{lcccccccccccccc}
        \toprule
        \multirow{2}{*}{\textbf{Methods}} &
        \multicolumn{4}{c}{\textbf{Text to Image}} &
        \multicolumn{3}{c}{\textbf{Depth to Image}} &
        \multicolumn{3}{c}{\textbf{Normal to Image}} &
        \multicolumn{2}{c}{\textbf{Depth Esti.}} &
        \multicolumn{2}{c}{\textbf{Normal Esti.}} \\
        \cmidrule(lr){2-5} \cmidrule(lr){6-8} \cmidrule(lr){9-11} \cmidrule(lr){12-13} \cmidrule(lr){14-15}
        & FID $\downarrow$ & CLIP $\uparrow$ & GD $\downarrow$ & GN $\downarrow$ & FID $\downarrow$ & CLIP $\uparrow$ & RMSE $\downarrow$ & FID $\downarrow$ & CLIP $\uparrow$ & RMSE $\downarrow$ & AbsRel$\downarrow$ & $\delta$1$\uparrow$ & m.$\downarrow$ & $11.25^\circ$$\uparrow$  \\
    \midrule
        JointNet Style& 21.71 & 30.53 & 7.82 & 8.24 &19.65 & 28.78 & 8.99 & 19.94  & 28.74  & 9.37 & 5.4 & 96.3 & 17.5 & 58.2 \\
        Marigold Style & 25.71 & 28.46 & 12.15 & 12.03 & 28.35 & 28.01 & 10.84 & 28.85  & 27.14  & 11.61 & 6.0 & 95.0 & 19.7 & 54.9 \\
        \textbf{\modelname~(Ours)} & \textbf{21.05} & \textbf{31.50} & \textbf{6.05} & \textbf{6.41} & \textbf{17.95} & \textbf{28.41} & \textbf{6.89} & \textbf{18.60} & \textbf{28.45} & \textbf{7.12} & \textbf{5.2} & \textbf{96.6} & \textbf{16.4} & \textbf{59.2} \\
    \bottomrule
    \end{tabular}}
\vspace{-6pt}
\end{table}

\noindent\textbf{Model Design Choices.}
As shown in~\cref{tab:ablation_design} and conceptualized in~\cref{fig:design}, we explored alternative designs on the SD3 backbone: duplicating the entire backbone (JointNet-style) and fully fine-tuning the backbone itself (Marigold-style). Importantly, even when utilizing the same powerful SD3 backbone, both alternative architectures yield significantly inferior performance compared to \methodname. This confirms that \modelname's superior performance stems primarily from our disentangled gradient flow and architectural design, rather than merely from scaling the backbone model.

\section{Conclusion and Limitations}
\label{sec:conclusion}

In this work, we present \modelname, an MMDiT-based framework for unified generation and perception tasks.
\modelname~comprises \methodname~and a unified dataset training strategy. The former uses only a copied image branch of MMDiT to model dense distributions beyond RGB, minimizing computational bloat, while the latter combines heterogeneous datasets into a unified framework to jointly optimize tasks efficiently.
\modelname~demonstrates outstanding performance across evaluations, surpassing existing unified models and performing on par with specialized regression expert models.
Crucially, our experiments reveal that generative priors and perceptual constraints provide complementary benefits, bridging the separated fields of diffusion-based synthesis and dense geometry prediction.

\noindent\textbf{Limitations.} While \modelname~provides a highly versatile framework, it inevitably introduces additional computational overhead. Specifically, duplicating the image branch increases the parameter count from 2B to 3B, and the 20-step inference time (FP16, NVIDIA V100) increases from 5.07s to 9.54s compared to the base model. However, considering that \modelname~effectively replaces multiple specialized network architectures with a single unified model capable of diverse controllable generation and dense prediction tasks, we consider this computational trade-off a reasonable and acceptable compromise for significantly expanded capabilities.

\begin{ack}
The research is supported in part by Early Career Scheme of the Research Grants Council (RGC) of the Hong Kong SAR under grant No. 26202321, ITF PRP/046/24FX, Science \& Technology Cooperation Program of Shandong under grant No. SDST26EG01, SAIL Research Project, and HKUST-Zeekr Collaborative Research Fund. This work was also supported by Damo Academy through Damo Academy Research Intern Program.
\end{ack}

\clearpage
\bibliographystyle{plain}
\bibliography{main}

\clearpage
\appendix

\section{Social Impacts}
\label{appendix:sec:limitation}
Our model equips the SOTA DiT with unified perception and generation capabilities but carries potential misuse risks, similar to other generative models. We emphasize the importance of responsible use and transparency.

\section{More Results}
\label{appendix:sec:results}

\subsection{Generalization to more perception tasks}
\begin{figure}[ht]
    \centering
    \includegraphics[width=0.99\linewidth]{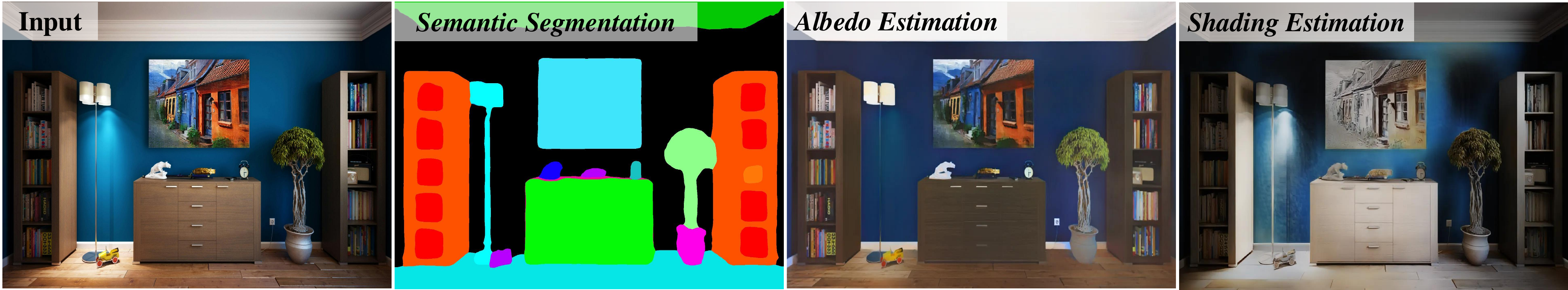}
    \caption{
    \modelname~can be adapted to more perception tasks.
    }
  \label{fig:more_perception}
\end{figure}

\modelname~builds on diffusion model, transforming vision tasks into conditional generation problems, enabling arbitrary perception tasks.
We trained \modelname~on additional perception tasks, including semantic segmentation, albedo estimation, and shading estimation, with qualitative results in \cref{fig:more_perception}.

\subsection{More qualitative results}
\label{supp:qualitative}
\begin{figure*}[http]
  \centering
  \includegraphics[width=0.95\linewidth]{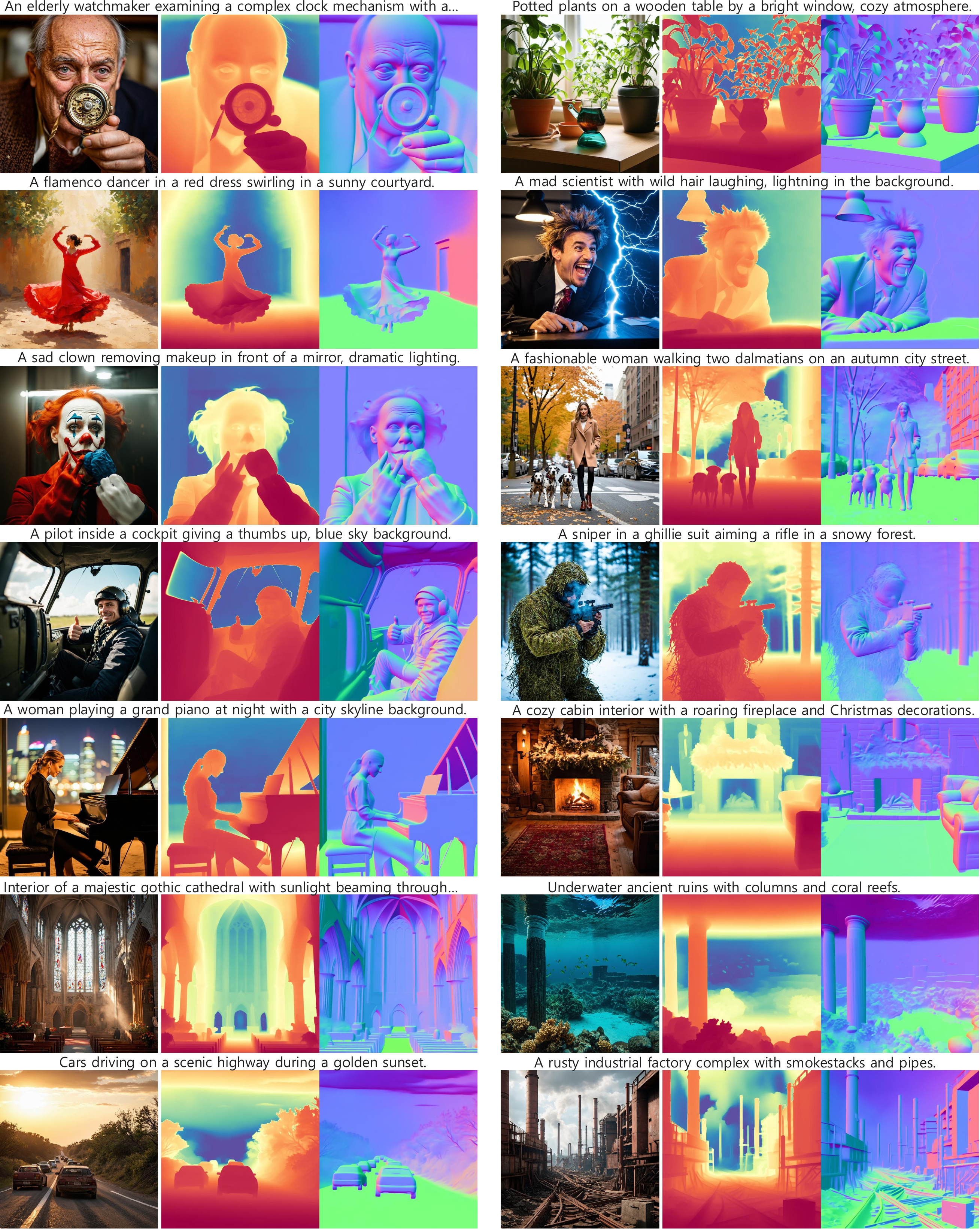}
  \caption{\textbf{Additional qualitative joint generation results}.
  }
  \label{fig:sup_qualitative_joint}
\end{figure*}

\begin{figure*}[http]
  \centering
  \includegraphics[width=0.95\linewidth]{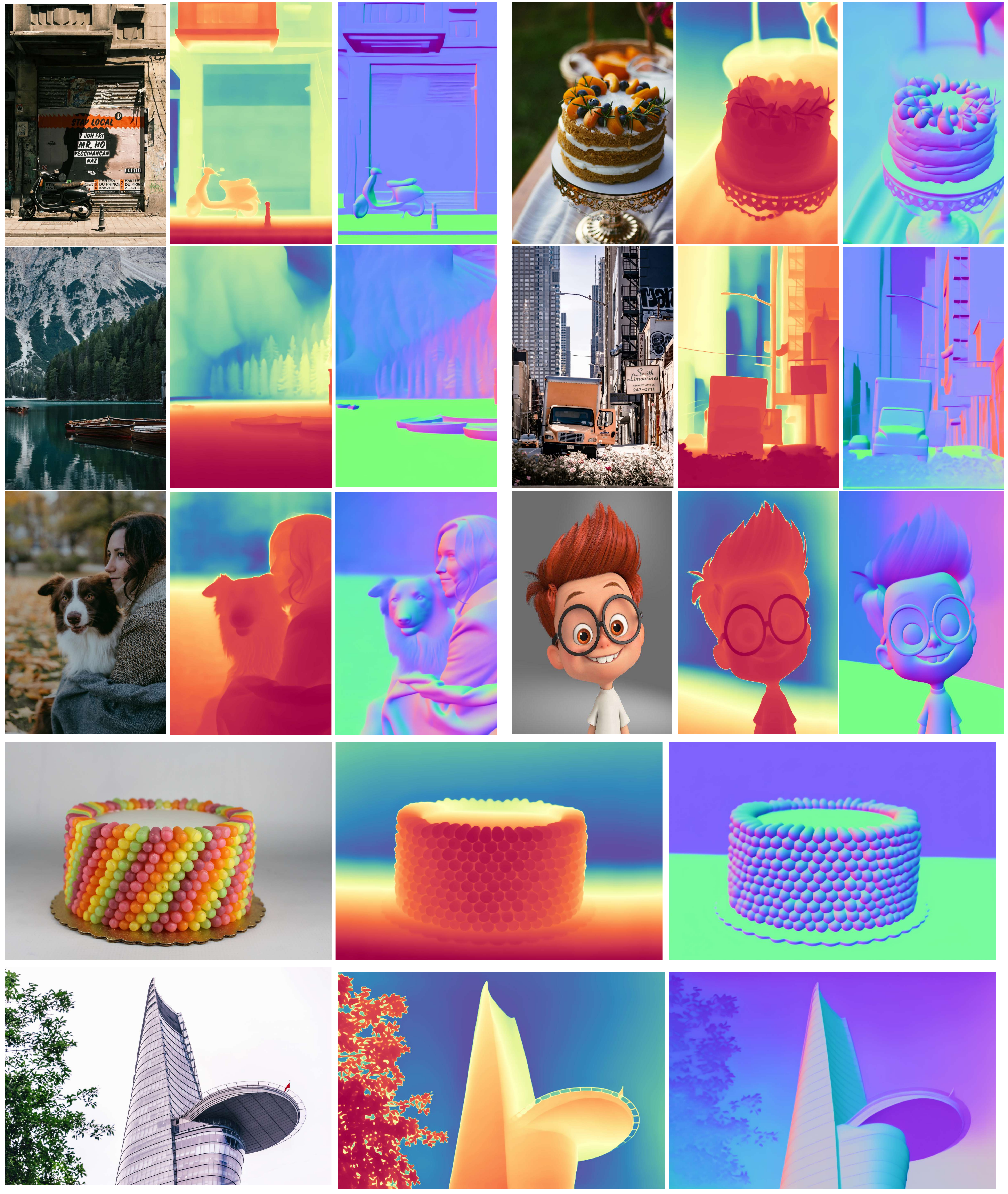}
  \caption{\textbf{Additional qualitative perception Results}.
  }
  \label{fig:sup_qualitative_1}
\end{figure*}

\begin{figure*}[http]
  \centering
  \includegraphics[width=0.95\linewidth]{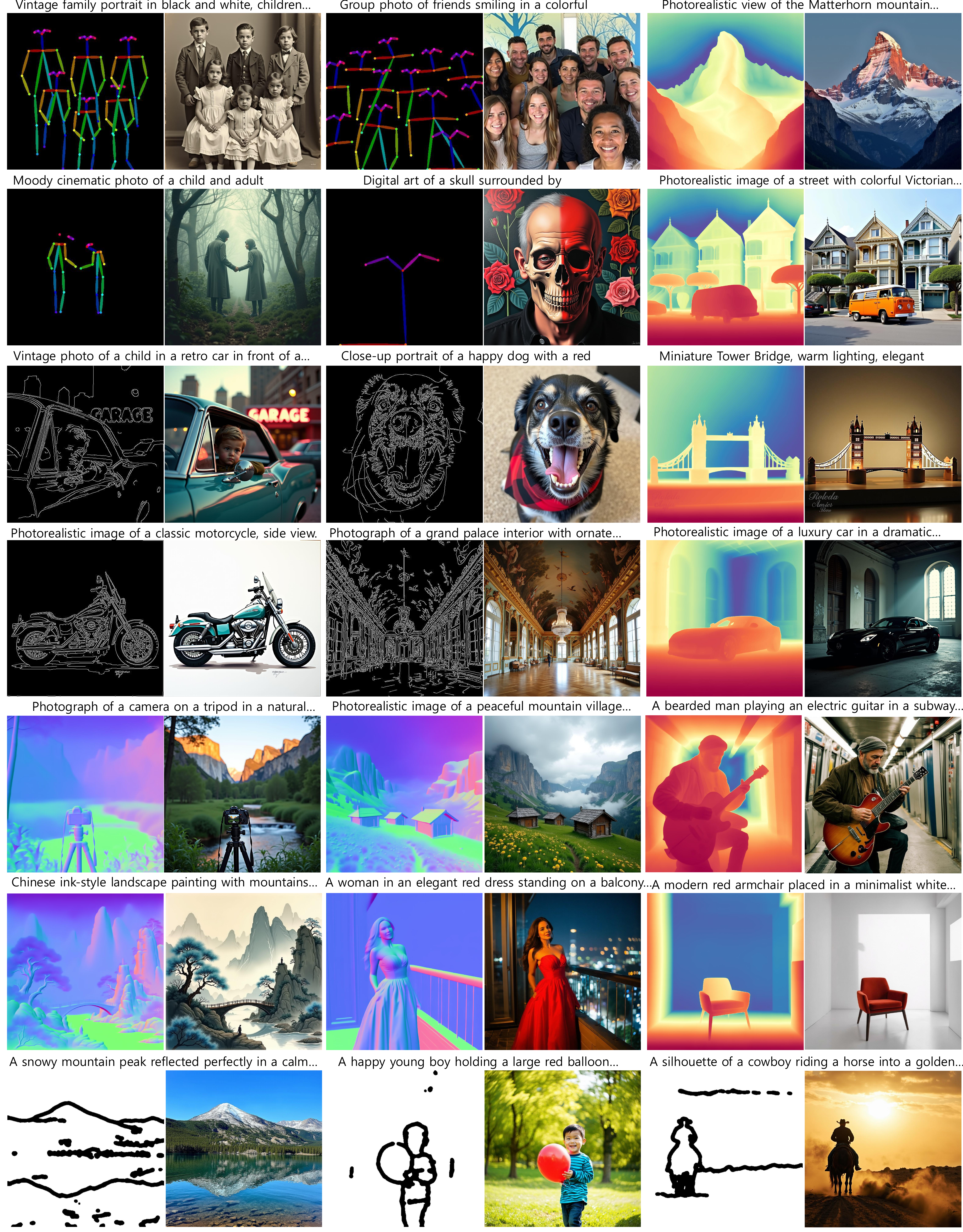}
  \caption{\textbf{Additional qualitative controllable generation results}.
  }
  \label{fig:sup_qualitative_2}
\end{figure*}

More qualitative results for joint generation are provided in \cref{fig:sup_qualitative_joint}. Additionally, qualitative examples for controllable generation and perception are shown in \cref{fig:sup_qualitative_1} and \cref{fig:sup_qualitative_2}, respectively.

\end{document}